\documentclass[runningheads]{llncs}
\usepackage{amsmath,amssymb}
\usepackage{graphicx}
\usepackage{booktabs}
\usepackage{orcidlink}
\usepackage{xcolor,colortbl}
\usepackage[nolist]{acronym}
\usepackage{subcaption}

\hypersetup{
    colorlinks,
    linkcolor={red!50!black},
    citecolor={blue!50!black},
    urlcolor={blue!80!black}
}
\definecolor{citecolor}{RGB}{34,139,34}
\definecolor{citecolor2}{HTML}{0071bc}
\definecolor{lightred}{RGB}{241,140,142}
\definecolor{defaultcolor}{gray}{0.9}
\definecolor{demphcolor}{gray}{.5}
\definecolor{myblue}{HTML}{659B90}
\definecolor{color2}{HTML}{005b46}
\definecolor{blockcolor}{HTML}{BFB69D}
\definecolor{blockcolor2}{HTML}{f8eed4}

\newcommand{\demph}[1]{\textcolor{demphcolor}{#1}}

\newcommand{\same}[1]{{\bf \fontsize{7.5}{42}\selectfont \color{black}~#1}}
\newcommand{\mrel}[1]{{\bf \fontsize{7.5}{42}\selectfont \color{lightred!180}~-#1}}
\newcommand{\prel}[1]{{\bf \fontsize{7.5}{42}\selectfont \color{citecolor!80}~+#1}}

\newcommand{\tablestyle}[2]{\setlength{\tabcolsep}{#1}\renewcommand{\arraystretch}{#2}\centering\small}
\newcommand{\tablestylesmall}[2]{\setlength{\tabcolsep}{#1}\renewcommand{\arraystretch}{#2}\centering\scriptsize}
\newcommand{\rrr}[1]{\raisebox{.3ex}{#1}}

\begin{acronym}[ECU]
\acro{mAP}{Mean Average Precision}
\acro{HOG}{Histogram of Oriented Gradients}
\acro{SAGHOG}[\textsc{Saghog}]{\textbf{S}elf-Supervised \textbf{A}utoencoder for \textbf{G}enerating \textbf{HOG} features}
\acro{ViT}{Vision Transformer}
\acro{CNN}{Convolutional Neural Network}
\acro{MAE}{Masked Autoencoder}
\acro{VLAD}{Vector of Locally Aggregated Descriptors}
\acro{ESVM}{Exemplar SVM}
\acro{SGR}{Similarity Graph Reranking}

\acro{WR}{Writer Retrieval}
\acro{SAM}{Segment Anything Model}
\end{acronym}

\begin{document}
\title{SAGHOG: Self-Supervised Autoencoder for Generating HOG Features for Writer Retrieval}
\titlerunning{\textsc{Saghog}}
%
\author{Marco Peer\orcidlink{0000-0001-6843-0830} \and Florian Kleber\orcidlink{0000-0001-8351-5066} \and
Robert Sablatnig\orcidlink{0000-0003-4195-1593}
}
\authorrunning{M. Peer et al.}
%
\institute{Computer Vision Lab, TU Wien\\
\email{\{mpeer, kleber, sab\}@cvl.tuwien.ac.at}\\
Code: \url{http://github.com/marco-peer/icdar24}\\
}
\maketitle              
\begin{abstract}
This paper introduces \textsc{Saghog}, a self-supervised pretraining strategy for writer retrieval using HOG features of the binarized input image. Our preprocessing involves the application of the Segment Anything technique to extract handwriting from various datasets, ending up with about 24k documents, followed by training a vision transformer on reconstructing masked patches of the handwriting.  \textsc{Saghog} is then finetuned by appending NetRVLAD as an encoding layer to the pretrained encoder. Evaluation of our approach on three historical datasets, Historical-WI, HisFrag20, and GRK-Papyri, demonstrates the effectiveness of \textsc{Saghog} for writer retrieval. Additionally, we provide ablation studies on our architecture and evaluate un- and supervised finetuning. Notably, on HisFrag20, \textsc{Saghog} outperforms related work with a mAP of 57.2 \% - a margin of 11.6~\% to the current state of the art, showcasing its robustness on challenging data, and is competitive on even small datasets, e.g. GRK-Papyri, where we achieve a Top-1 accuracy of 58.0~\%.
\end{abstract}
\section{Introduction}
\ac{WR} is the task of locating documents written by the same author within a dataset, achieved by identifying similarities in handwriting \cite{keglevic}. This task is particularly valuable for historians and paleographers, allowing them to track individuals or social groups across various historical periods \cite{icdar19}. Additionally, \ac{WR} is used for recognizing documents with unknown authors and uncovering similarities within such documents \cite{unsupervised_icdar17}.

\ac{WR} methods usually consist of multiple steps, such as sampling of handwriting by applying interest point detection, a neural network for feature extraction, and feature encoding such as NetVLAD, followed by aggregation of the encoded features \cite{peer_netrvlad}. Although those methods work for large datasets such as Historical-WI \cite{icdar2017}, \ac{WR} still lacks performance for datasets that contain less handwriting or noise such as degradation, best exemplified by HisFrag20 \cite{hisfrag20}. While approaches trained on full fragments currently work best \cite{avlad}, experiments show that those methods do infer features from the background, not necessarily related on the actual handwriting \cite{peer_fm}.

\begin{figure}[t]
    \centering
     \includegraphics[width=\textwidth]{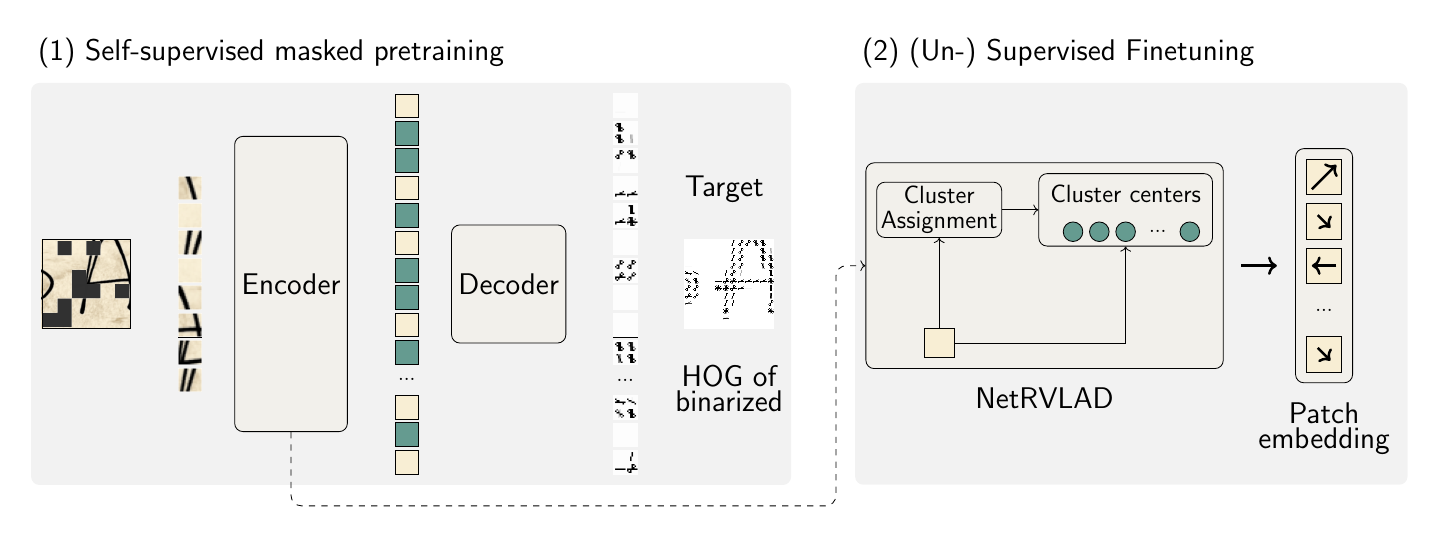}
    \caption{Overview of \ac{SAGHOG}. The decoder reconstructs the HOG features of the masked tokens ( \rrr{\fcolorbox{black}{myblue}{\rule{0pt}{3pt}\rule{3pt}{0pt}}} ) by only using the non-masked tokens ( \rrr{\fcolorbox{black}{blockcolor2}{\rule{0pt}{3pt}\rule{3pt}{0pt}}} ). In the second stage, the pretrained encoder with NetRVLAD is finetuned, either by training on writer or pseudo labels.}
    \label{fig:enter-label}
\end{figure}

With the recent gain of interest in self-supervised algorithms that make use of large amounts of unlabeled data, this paper introduces \ac{SAGHOG}, a self-supervised pretraining strategy based on \ac{HOG} features. \ac{SAGHOG} is two-staged and relies on the \ac{ViT} architecture, as shown in Fig.~\ref{fig:enter-label}: Firstly, we aim to close that gap of performance for complex datasets by self-supervised pretraining on reconstructing only \ac{HOG} features of the handwriting. Since related work relies on clustering SIFT descriptors \cite{unsupervised_icdar17,peer_netrvlad}, predicting \ac{HOG} features is an intuitive choice. Secondly, we add NetRVLAD \cite{peer_netrvlad} to the pretrained encoder and finetune the model on the respective dataset. We apply and evaluate two finetuning strategies: supervised - using the writer label - or unsupervised finetuning - using the cluster of the corresponding SIFT descriptor of the input patch as a surrogate class (Cl-S) \cite{unsupervised_icdar17}. Additionally, as a preprocessing step, we propose to apply \ac{SAM} \cite{sam}, a foundation model for segmentation, to extract the handwriting of the data, accelerate the training step, and improve the generalization ability of \ac{SAGHOG}.

We evaluate our results on three datasets: Historical-WI \cite{icdar2017}, HisFrag20 \cite{hisfrag20} and GRK-Papyri \cite{grk-papyri}, each corresponding to a different domain. We provide ablation studies on our preprocessing stage, different target features of \ac{SAGHOG} and hyperparameters of the pretraining stage and the network architecture. Our experiments show that \ac{HOG} reconstruction during pretraining does improve \ac{WR} when finetuning with NetRVLAD, in particular on HisFrag20, where we outperform the state of the art by a large margin (11.6~\%). Furthermore, we find that only optimizing NetRVLAD is able to outperform finetuning the network, showing that \ac{SAGHOG} is beneficial for \ac{WR}. On Historical-WI, the effect of pretraining is limited, but using \ac{SAGHOG} does not harm the performance and is competitive with other approaches. In the end, we show that even on small, out-of-domain datasets such as GRK-Papyri, \ac{SAGHOG} shines, in particular in terms of Top-1 accuracy, making \acp{ViT} valuable for further research in the domain of \ac{WR}. 

Concluding, our contributions are summarized to:
\begin{itemize}
    \item We collect multiple datasets and propose a data curation scheme to enable pretraining of \ac{SAGHOG}.
    \item \ac{SAGHOG} is the first self-supervised masking-based approach for handwriting of $32\times32$ patches without the need for any carefully designed data augmentation pipeline. We investigate two different finetuning schemes - supervised and unsupervised.
    \item We thorougly evaluate our approach on Historical-WI \cite{icdar2017}, HisFrag20 \cite{hisfrag20} and GRK-Papyri \cite{grk-papyri} and outperform state-of-the-art methods on \ac{WR} as a downstream task.
\end{itemize}

 Our code is publicly available. The remainder of our paper is structured as follows. Section~\ref{sec:rel_work} gives details on related work in the field of self-supervised learning and \ac{WR}. Afterwards, we describe \ac{SAGHOG} in Section~\ref{sec:method}, including pretraining and finetuning, followed by our evaluation protocol in Section~\ref{sec:eval} and our results in Section~\ref{sec:results}. We summarize our findings in Section~\ref{sec:conclusion}.

\section{Related Work}\label{sec:rel_work}

In this section, we briefly discuss related work. We start with the developments in self-supervised learning with a focus on \ac{WR}, followed by the current state-of-the-art \ac{WR} methods.

\paragraph{Self-Supervised Learning}  In recent years, self-supervised learning has gained popularity to learn meaningful representations from unlabeled data. While metric learning approaches like SimCLR \cite{simclr} or MoCo \cite{moco} involve training two separate networks to generate similar embeddings for a sample, they are outperformed by self-distillation methods, e.g. DINO \cite{dino}, that use a carefully designed augmentation pipeline to generate different views of an image. In the domain of \ac{WR}, we introduced a self-supervised algorithm based on DINO \cite{dino} using morphological operations to generate augmented views of the binarized Historical-WI dataset \cite{peer_selfsupervised}. Lastilla et al. \cite{Lastilla2022} apply an online bag-of-words pretraining to documents of the Vatican Apostolic Library for \ac{WR}. Both approaches improve the \ac{mAP} on the datasets evaluated, but they use large crops which slows down training. Additionally, they do not use any encoding or finetuning at all. However, those augmentation pipelines suffer when training on small patches only containing a few strokes of handwriting. Therefore, we focus on another approach of self-supervision called masked image modeling or \ac{MAE}, introduced by He et al. \cite{mae}, where a network learns to reconstruct masked parts of an image. An intuitive adaption of \ac{MAE} is MaskFeat \cite{maskfeat}, proposed by Wei et al., where the target features are handcrafted descriptors rather than pixel values, e.g. \ac{HOG}. \ac{SAGHOG} also applies this strategy by adapting MaskFeat to the binarized version of the input image. This is an intuitive choice, since dominating approaches for \ac{WR} rely on SIFT descriptors \cite{unsupervised_icdar17,peer_netrvlad} for training, and those descriptors are formed by aggregating \ac{HOG} features. 

\paragraph{Writer Retrieval} \ac{WR} methods fall into two categories: codebook-based and codebook-free approaches. Codebooks serve as models to compute handwriting statistics, with \ac{VLAD} standing out as a prominent method for \ac{WR} \cite{zernike,unsupervised_icdar17,christlein_cnn_vlad,christlein_papyri}. Handwriting characteristics are extracted either through handcrafted features \cite{zernike,bVLAD} or CNNs \cite{unsupervised_icdar17,christlein_cnn_vlad,christlein_papyri}. For historical datasets in particular, Christlein et al. \cite{unsupervised_icdar17} demonstrate that training on $32\times32$ patches with pseudolabels generated by clustering SIFT descriptors outperforms supervised methods \cite{wang_supervised}. Additionally, \ac{ESVM} refines each descriptor's encoding by training multiple SVMs with only one positive sample and the remaining ones as negatives. The HisIR19 competition winners \cite{icdar19} and the current state-of-the-art on Historical-WI dataset utilize handcrafted features (SIFT and pathlet) for retrieval, encoding them via bagged VLAD (bVLAD) \cite{bVLAD}. In our previous work \cite{peer_netrvlad} we employ NetRVLAD, a NetVLAD-based layer with reduced complexity, to train the neural network in and end-to-end way. Additionally, we introduced \ac{SGR}, a reranking strategy that shows superior performance compared with state of the art. In contrast to training with small patches, on more complex datasets including fragments, such as HisFrag20 \cite{hisfrag20}, leading approach train on full images: Ngo et al. \cite{avlad} propose an attention-based \ac{VLAD} called A-VLAD, while our feature mixer \cite{peer_fm} uses an codebook-free encoding stage based on convolutional and fully connected layers. In this work, finetuning of \ac{SAGHOG} is closely related to our training protocol of NetRVLAD \cite{peer_netrvlad}, and we show that our approach is outperforming methods on image-level for datasets such as HisFrag20. \ac{SAGHOG} is also based on the \ac{ViT} architecture instead of relying on CNNs.

\section{Methodology}\label{sec:method}
In this section, we cover each part of our proposed approach. We start with the curation of the dataset for pretraining, followed by describing \ac{SAGHOG}. In the end, NetRVLAD finetuning and feature aggregation are explained.

\subsection{Data Curation}

\begin{figure}[t!]
     \centering
              \begin{subfigure}[b]{0.6\textwidth}
          \includegraphics[width=\textwidth]{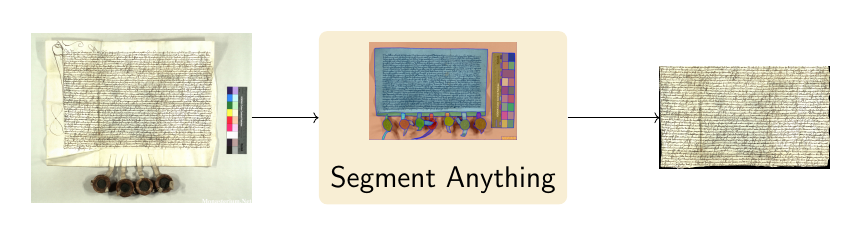}
         \caption{Handwriting extraction via SAM}\label{fig:sam}
         \end{subfigure}
         
         \begin{subfigure}[b]{0.6\textwidth}
         \centering

            \medskip
                 \includegraphics[width=\textwidth]{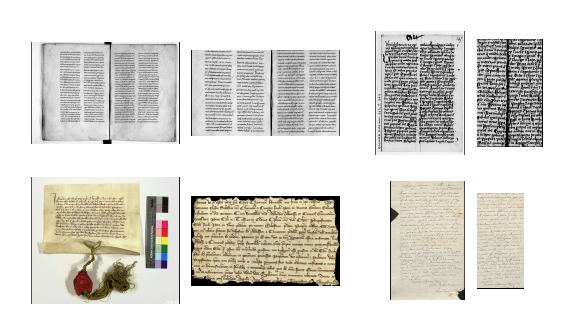}

                      \caption{Examples of input-output pairs}\label{fig:qual_examples}
     \end{subfigure}
              \begin{subfigure}[b]{0.35\textwidth}
         \centering
     \includegraphics[width=\textwidth]{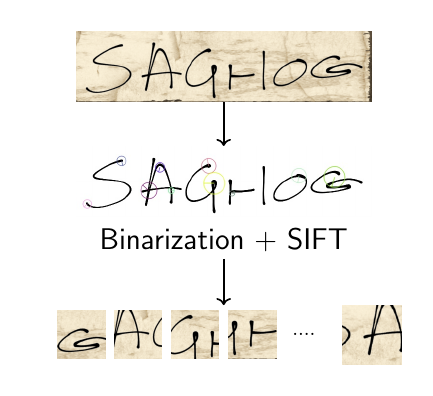}
         \caption{Patch sampling} \label{fig:sampling}
     \end{subfigure}
    \caption{Preprocessing scheme. (\subref{fig:sam}) We obtain the handwriting of a page by processing the segmentation of \ac{SAM}. (\subref{fig:qual_examples}) Examples of the produced images. (\subref{fig:sampling}) $32\times 32$ patches are extracted by applying SIFT on the binarized image.}
    \label{fig:preprocessing}
\end{figure}

Self-supervised learning relies on a large amount of (unlabelled) data. Hence, we employ multiple datasets proposed in previous competitions in the domain of historical document analysis for our pretraining. For the datasets used, presented in Table~\ref{tab:datasets}, we are mainly inspired by the choice of Zenk et al. \cite{Zenk2023}. Our preprocessing scheme, also shown in Fig.~\ref{fig:preprocessing}, is two-fold: \emph{handwriting extraction} and \emph{patch sampling}.

\begin{table}
\caption{Datasets used for self-supervised pretraining after preprocessing.}\label{tab:datasets}
\vspace{5pt}
\tablestyle{5.5pt}{1.05}
\begin{tabular}{lll}\toprule
Dataset &  Subset & Images \\
\midrule
CLaMM 2016~{\scriptsize\cite{clamm2016}} & Task 1 / Task 2 & 2607 \\
Historical-WI~{\scriptsize\cite{icdar2017}} & Train (color) & 1182 \\
CLaMM 2017~{\scriptsize\cite{clamm2017}} & Task 1 / Task 3 & 1572 \\
cBAD 2019~{\scriptsize\cite{cbad}} & Validation / Test & 531 \\
HisIR19~{\scriptsize\cite{icdar19}} & Validation / Test & 1152 / 17641 \\ \midrule
Total & ~ & 24685 \\
\bottomrule
\end{tabular}
\end{table}

\emph{Handwriting extraction} Since datasets such as HisIR19 \cite{icdar19} contain additional elements like color palettes, book covers, and background noise, we use \ac{SAM} \cite{sam}, a foundation model for segmentation. We use the default pretrained model and process the data as follows:
\begin{enumerate}    
    \item We apply a Canny edge detector on the image.
    \item Since \ac{SAM} relies on a grid of points and we only want to obtain the handwriting part that is usually a significant part of the page, we use a small $8\times8$ grid for segmentation.
    \item To discard wrong results, we filter masks with less than 10~\% edge pixels and a confidence level lower than 0.8.
\end{enumerate}

If the preceding steps generate multiple masks, we merge them, restricting the maximum number of masks per page to two. Additionally, to address pages with minimal handwriting or none at all, we discard pages with fewer than 1000 SIFT keypoints detected. This procedure allows us to take smaller random crops of the pages to extract keypoints, which significantly accelerates the processing time during training and removes non-handwriting areas. As depicted in Fig.~\ref{fig:qual_examples}, we occasionally oversegment and remove some handwriting. Note that we do not apply \ac{SAM} on the trainset of Historical-WI \cite{icdar2017} since it contains pure handwriting. To show the effectiveness of the preprocessing, we annotate handwriting areas of 200 pages of the HisIR19 \cite{icdar19} test set and evaluate the bounding boxes. We achieve an IoU@0.5 of 74~\%. This is mostly due to \ac{SAM} splitting the handwriting areas in multiple zones (see Fig.~\ref{fig:qual_examples}). However, all of our detected masks are inside the annotated handwriting areas, indicating that we can successfully segment the handwriting - although we remove parts of it.

\emph{Patch sampling} During training, we take a random crop of size 256 and sample small parts of the handwriting by extracting $32\times 32$ patches located around SIFT keypoints. The keypoints are detected on the binarized version of the page via SIFT, and we discard patches with less than 1~\% black pixels, similar to previous work \cite{unsupervised_icdar17,peer_netrvlad}. We investigate the influence of the binarization in our evaluation in Section~\ref{sec:results}.

The composition of our final dataset used for pretraining is shown in Table~\ref{tab:datasets}. Although the majority of images are included in the HisIR19~\cite{icdar19} dataset, it contains various document types such as manuscript books, letters, or charters.

\subsection{Self-Supervised Pretraining}

\ac{SAGHOG} is based on the \ac{MAE} architecture  \cite{mae} consisting of an encoder-decoder structure. We use the vanilla \ac{ViT} \cite{vit} structure with a patch size of four. Firstly, random tokens are masked out - by default, we mask 75~\% of the image. The encoder projects each unmasked token of the $32\times32$ input color image to an embedding of size 512. Those embeddings are then projected by the decoder to the output feature space: While \ac{MAE} \cite{mae} and MaskFeat \cite{maskfeat} use RGB pixel values or \ac{HOG} features of the colorized input, we propose to reconstruct the \ac{HOG} features of the binarized image due to two reasons. By relying on binarization, the network automatically learns to focus on handwriting, and on the other hand, we are able to suppress the influence of the background. \ac{HOG} features are calculated by applying a convolutional layer with fixed weights to obtain the gradients. Afterward, we build a 9-bin histogram on cells of size $4\times 4$, resulting in a feature vector of dimension 9 for each token. The final \ac{HOG} features are $l_2$ normalized.

\begin{figure}
    \centering
    \begin{subfigure}[b]{0.2\textwidth}
        \includegraphics[width=\textwidth]{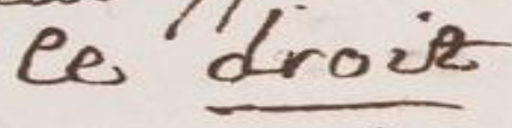}
        \includegraphics[width=\textwidth]{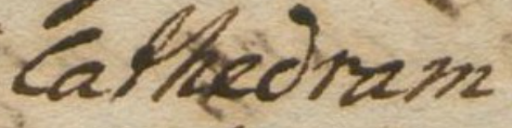}
        \includegraphics[width=\textwidth]{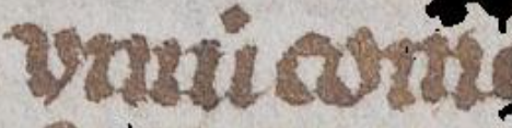}
        \includegraphics[width=\textwidth]{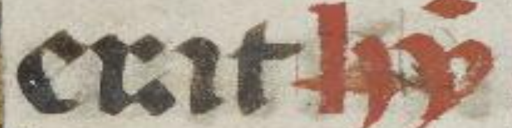}
        \caption{}\label{fig:hog_inp}
    \end{subfigure}\hfill
        \begin{subfigure}[b]{0.2\textwidth}
        \includegraphics[width=\textwidth]{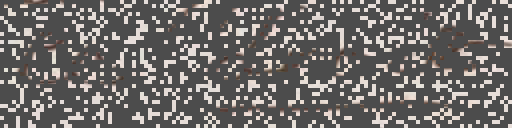}
        \includegraphics[width=\textwidth]{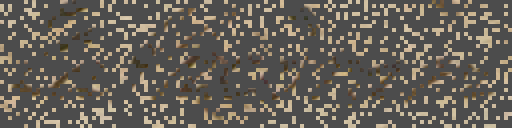}
        \includegraphics[width=\textwidth]{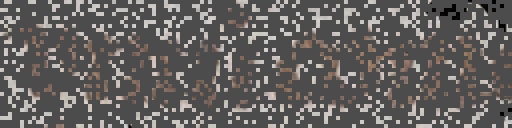}
        \includegraphics[width=\textwidth]{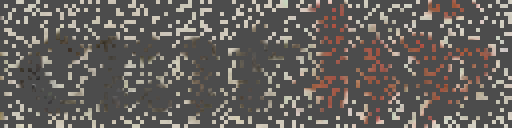}
        \caption{}\label{fig:hog_masked}
    \end{subfigure}\hfill
        \begin{subfigure}[b]{0.2\textwidth}
        \includegraphics[width=\textwidth]{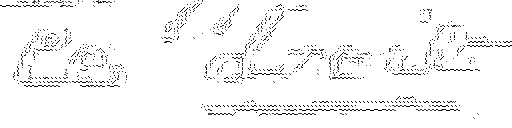}
        \includegraphics[width=\textwidth]{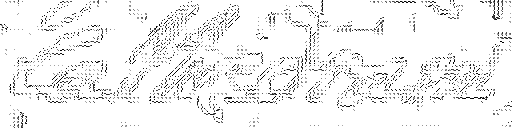}
        \includegraphics[width=\textwidth]{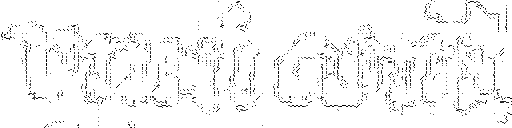}
        \includegraphics[width=\textwidth]{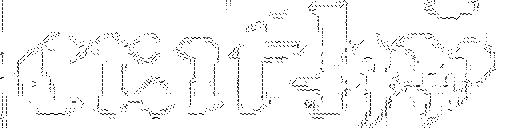}
        \caption{}\label{fig:hog_pred}
    \end{subfigure}\hfill
        \begin{subfigure}[b]{0.2\textwidth}
        \includegraphics[width=\textwidth]{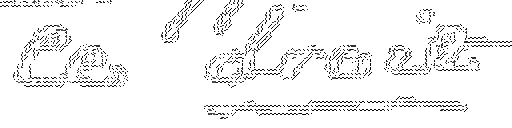}
        \includegraphics[width=\textwidth]{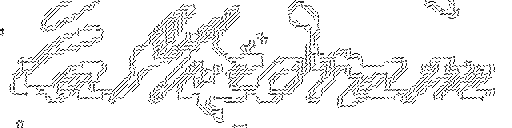}
        \includegraphics[width=\textwidth]{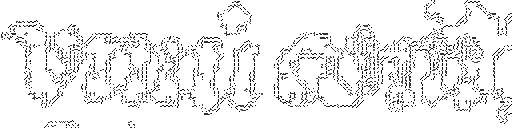}
        \includegraphics[width=\textwidth]{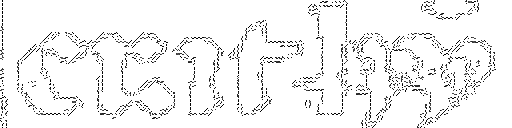}
        \caption{}\label{fig:hog_gt}
    \end{subfigure}
    \caption{Examples of reconstructed \ac{HOG} features. For better visualization, we stack multiple patches to create images of (128, 512). The images are not seen during pretraining. (\subref{fig:hog_inp}) Raw input image. (\subref{fig:hog_masked}) Masked input with a ratio of 0.75. (\subref{fig:hog_pred}) Prediction of the missing \ac{HOG} features of the binarized image. (\subref{fig:hog_gt}) Actual \ac{HOG} features.}
    \label{fig:hog}
\end{figure}

We choose, similar to related work \cite{maskfeat}, a larger encoder consisting of 8 layers (depth) and a smaller decoder (one layer) since this shows improved performance on the downstream task. We denote the encoder as \ac{ViT}-4/8.  Qualitative examples of the training process are depicted in Fig.~\ref{fig:hog}. In general, the network is able to reconstruct the missing tokens (and therefore the style of handwriting as well as parts of letters), even if only small parts of the handwriting are visible. We evaluate the choice of target features as well as the hyperparameters of \ac{SAGHOG} in our evaluation.

\subsection{Finetuning}
After pretraining of \ac{SAGHOG}, we drop the decoder and only use the encoder for further finetuning. We use the class token as output of the encoder, a learnable embedding that describes the global information of the token sequence. We also studied pooling of all patch embeddings instead of using the class token, but this did not yield significant differences. Finetuning is then performed by appending the NetRVLAD layer \cite{peer_fm} with 100 cluster centers and a final $l_2$-normalization of the encoded embedding. In our evaluation, we investigate two finetuning schemes commonly used in the domain of \ac{WR}:

\begin{enumerate}
\item \emph{Supervised}: The target label is the writer of the document, this refers to the default supervised finetuning strategy. 
\item \emph{Cl-S}: The target label is the cluster membership of the clustered SIFT descriptors extracted at the corresponding keypoints. We follow the protocol used in \cite{peer_netrvlad}: We cluster the trainset by using k-Means with 5000 centers and only consider patches that pass the ratio test between the distance to the nearest and the second nearest cluster ($< 0.9$). This has shown to perform better than supervised training for \ac{WR} \cite{unsupervised_icdar17,peer_netrvlad}. 
\end{enumerate}

\subsection{Writer Retrieval}

We aggregate all patch embeddings $\boldsymbol{\mathrm{x}}_i$ with $i = 0, \dots, n_{p} -1$ of a page using $l_2$ normalization followed by sum pooling $\boldsymbol{\mathrm{X}} = \sum_{i=0}^{n_p-1} \boldsymbol{\mathrm{x}}_i$ to obtain the global page descriptor $\boldsymbol{\mathrm{X}}$. Furthermore, to reduce visual burstiness \cite{visual_burstiness}, we apply power-normalization $f(x) = \text{sign}(x)|x|^\alpha$ with $\alpha = 0.4$, followed by $l_2$-normalization. Finally, a dimensionality reduction with whitening via PCA is performed. 

\ac{WR} is then evaluated by a \emph{leave-one-out strategy}: Each image of the test set is once used as a query $q$, and the remaining documents are ranked according to the cosine similarity of the global page descriptors.

\section{Evaluation}\label{sec:eval}
We provide details on the datasets used for evaluation and our experimental setup in this section.
\subsection{Datasets}
\setlength{\fboxsep}{0pt}
\begin{figure}
    \centering
    \begin{subfigure}[b]{0.3\textwidth}\centering
        \fbox{\includegraphics[height=1.1\textwidth]{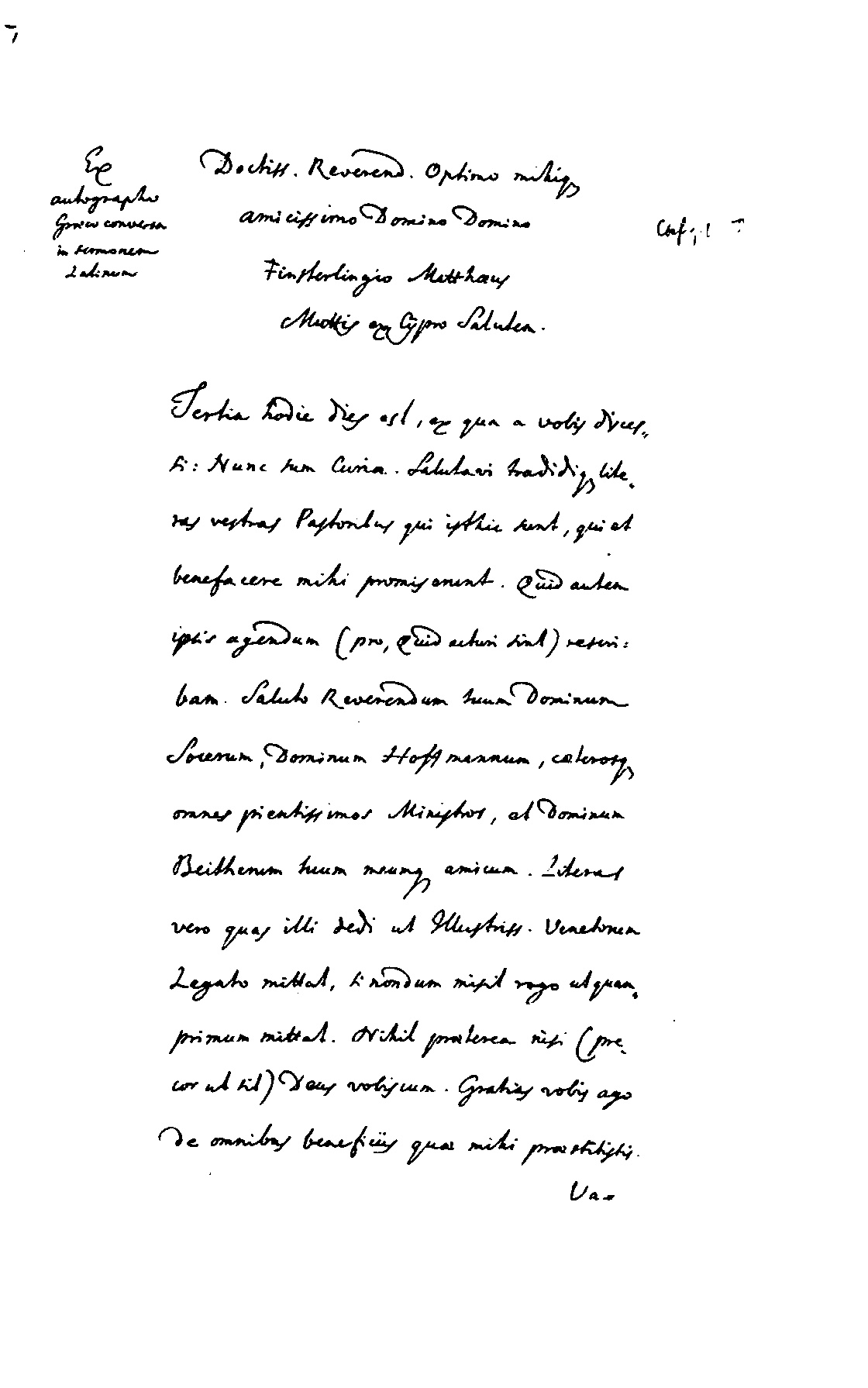}}
         \caption{Historical-WI}\label{fig:dataset_icdar2017}
     \end{subfigure}
    \begin{subfigure}[b]{0.3\textwidth}\centering
         \fbox{\includegraphics[width=0.7\textwidth]{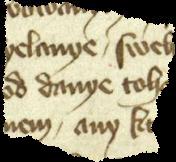}}
         \fbox{\includegraphics[width=0.7\textwidth]{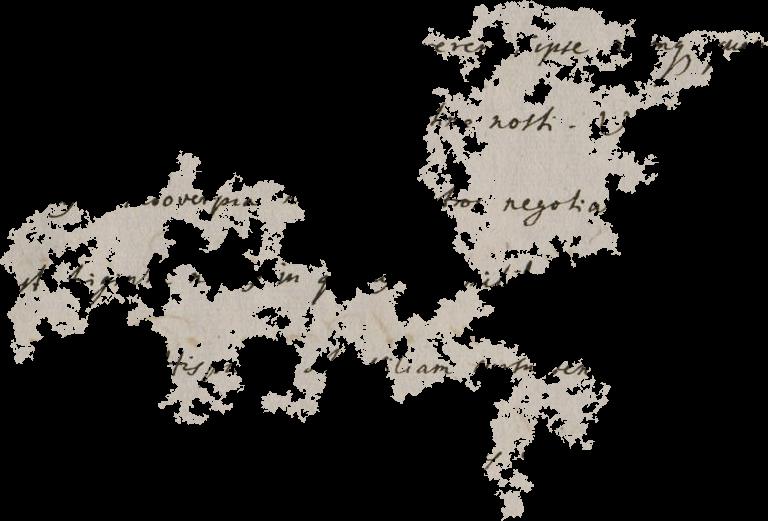}}
         \caption{HisFrag20}\label{fig:dataset_hisfrag20}
         \end{subfigure}
    \begin{subfigure}[b]{0.33\textwidth}\centering
         \fbox{\includegraphics[width=\textwidth]{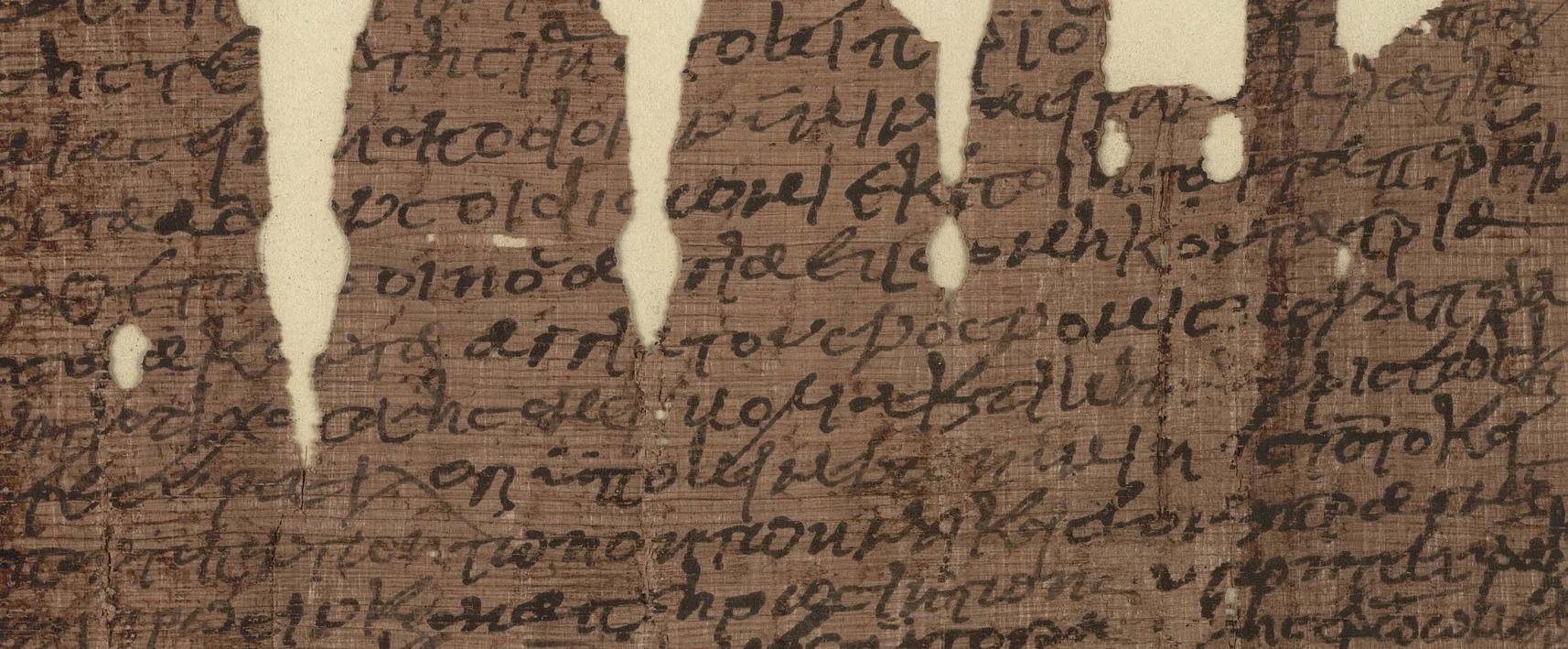}}\\
         \fbox{\includegraphics[width=\textwidth]{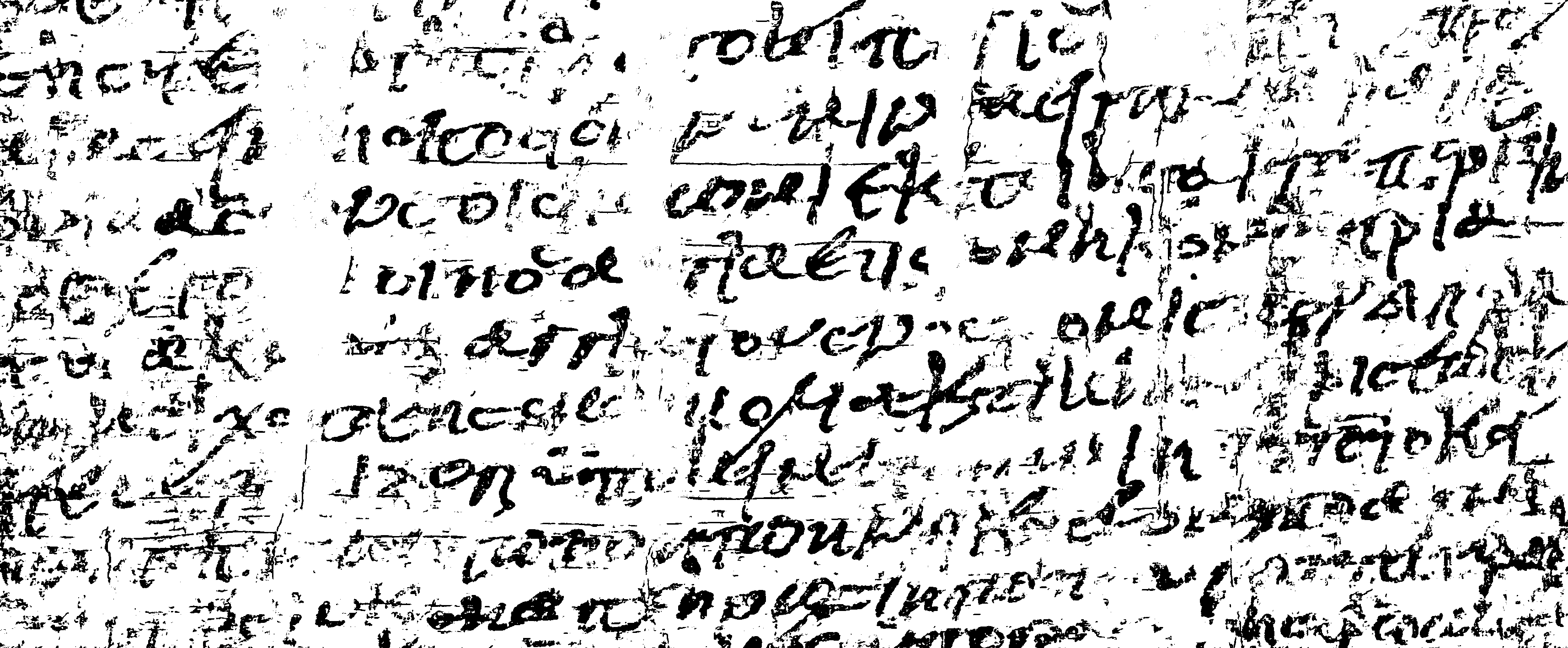}}
         \caption{GRK-Papyri}\label{fig:dataset_grk}
     \end{subfigure}
     \caption{Examples of the datasets used. (\subref{fig:dataset_icdar2017}) Binarized example of Historical-WI testset. (\subref{fig:dataset_hisfrag20}) HisFrag20: Sample of train- and testset. (\subref{fig:dataset_grk}) GRK-Papyri (Dioscorus-5) color and binarized version \cite{christlein_papyri} - we only use the binarized set.}
    \label{fig:datasets}
\end{figure}

We evaluate our approach on three datasets, as shown in Fig.~\ref{fig:datasets}, each representing a different scenario:

\paragraph{Historical-WI} Introduced by Fiel et al. \cite{icdar2017}, the dataset consists of 3600 images and 760 writers, each contributing five pages. We use the binarized version for evaluation, following related work. The training set includes 1192 pages. We consider Historical-WI as a dataset of moderate size and since it is binarized without containing any additional noise, handcrafted features \cite{bVLAD} as well as small networks such as ResNet20 \cite{unsupervised_icdar17,peer_netrvlad} work reasonably well.
\paragraph{HisFrag20} The main dataset used in our work is the HisFrag20 \cite{hisfrag20}, introduced via the \emph{ICFHR 2020 Competition on Image Retrieval for Historical Handwritten Fragments}. The trainset consists of about 100k fragments; the respective test set includes 20019 fragments from 1152 writers and 2732 documents. The generation algorithm for the fragments differs between the train- and test set, making the dataset challenging, which results in a lack of performance compared to Historical-WI.
\paragraph{GRK-Papyri} Additionally, we provide baseline results for comparison on GRK-Papyri \cite{grk-papyri}, a small dataset consisting of only 50 documents of 10 different writers.  Note that this dataset represents a heavy domain shift compared to the datasets used for pretraining - from Latin to ancient Greek, as well as high degradation, e.g., holes, stains, and low contrast. To compare to related work, we use the U-Net binarized version provided by Christlein et al. \cite{christlein_papyri}. Even though binarization tackles most of the degradation, binarized images still suffer from noise and missing strokes.

\subsection{Experimental Setup}
In the following, the setups for both training stages as well as for the aggregation step are given.

\paragraph{Pretraining} We use the vanilla \ac{ViT} architecture with a patch size of 4 and a depth of 8, denoted as ViT-4/8. The cell size for calculating \ac{HOG} features is $4\times 4$, and the decoder depth is 1, as related work \cite{mae,maskfeat} shows negligible influence of those parameters. During training, we take a random crop of size $256\times256$ and apply grayscale and binarization (both with a probability of $p=0.2$). One batch consists of 64 pages, each contributing 32 patches, yielding a batch size of 2048. Further hyperparameters are given in Table~\ref{tab:pretrain_hyp}.

\begin{table}
\caption{Hyperparameters of our training, if not stated otherwise.}
\centering
\begin{subtable}[t]{0.32\textwidth}
\centering
        \caption{Pretraining}\label{tab:pretrain_hyp}
        \tablestylesmall{5pt}{1.05}
        \begin{tabular}{ll}
        \toprule
        Config  & ViT-4/8  \\ \midrule
        Optimizer & AdamW \cite{adamw} \\
        Weight decay & 0.05 \\
        Learning rate & $8\cdot 10^{-4}$ \\
        Scheduler & Cosine decay \cite{cosine_decay}  \\
        Batch size & 2048 \\
        Gradient clipping & 0.02\\
        Warmup epochs &  5\\
        Epochs & 200  \\
        \bottomrule
        \end{tabular}
    \end{subtable}%
    \hfill
\begin{subtable}[t]{0.64\textwidth}
\centering
        \caption{Finetuning}\label{tab:finetune_tab}
        \tablestylesmall{5pt}{1.05}
        \begin{tabular}{lcc}
        \toprule
        Config  & ViT-4/8 & ResNet \\ \midrule
        Optimizer & AdamW \cite{adamw} & Adam \cite{adam} \\
        Weight decay & 0.01 & 0 \\
        Learning rate & \multicolumn{2}{c}{$10^{-3}$} \\
        Scheduler & \multicolumn{2}{c}{Cosine decay \cite{cosine_decay}} \\
        Batch size & \multicolumn{2}{c}{1024}\\
        Gradient clipping & \multicolumn{2}{c}{1.0} \\
        Warmup epochs & \multicolumn{2}{c}{5} \\
        Epochs & \multicolumn{2}{c}{50} \\
        \bottomrule
        \end{tabular}
    \end{subtable} 
\end{table}

\paragraph{Finetuning} We finetune all networks by appending NetRVLAD \cite{peer_netrvlad} with 100 cluster centers to \ac{SAGHOG}. The training is done for 50 epochs on triplets with Multi-Similarity loss \cite{msloss} and a batch size of 1024 with 16 samples per class. For data augmentation, we apply erosion and dilation with a randomly sampled $3\times3$ kernel and a probability of 0.1. Additionally, token dropout is applied ($p=0.1$). We use an early stopping of five epochs if the \ac{mAP} on the validation set does not increase. All of our networks reach convergence with these settings (10~\% of the writers are used as validation). Further hyperparameters are given in Table~\ref{tab:finetune_tab}.

\paragraph{Retrieval} If not stated otherwise, we report the \ac{mAP} on the task of \ac{WR} for the respective datasets when \ac{SAGHOG} is jointly finetuned with NetRVLAD in an unsupervised manner (Cl-S). The Top-1 metric indicates if the first document of the ranked list is a hit. The retrieval is performed on PCA-whitened global descriptors of dimension 512. 

\section{Results}\label{sec:results}
In this section, we evaluate different parts of our pipeline and study the gain of performance by (un-)supervised finetuning of \ac{SAGHOG}.

\subsection{Ablation studies}

\paragraph{Preprocessing} Firstly, in Table~\ref{tab:sam_preprocessing}, we study the influence of preprocessing via \ac{SAM}. For both datasets, \ac{SAGHOG} performs better when we preprocess data for pretraining with \ac{SAM}, in particular on the HisFrag20 dataset, where we report a gain of 4.1~\% \ac{mAP}. Note that pretraining with no preprocessing includes 27k images instead of 24k due to the filtering applied after using \ac{SAM}.

\begin{table}
\caption{\ac{SAM} preprocessing.}\label{tab:sam_preprocessing}        
    \tablestyle{5pt}{1.05}
        \begin{tabular}{lcc}
        \toprule
        ~  & Historical-WI & HisFrag20 \\ \midrule
        \demph{From scratch} & \demph{71.1} & \demph{47.4} \\ \midrule
        Full image & 72.8 & 50.5 \\
        With \ac{SAM} & \textbf{73.3} &  \textbf{54.6}  \\ 
        \bottomrule
        \end{tabular}
    \end{table}%

\paragraph{Target features} Next, we evaluate the choice of the target features of the decoder. We train four networks with different features, as shown in Table~\ref{tab:tar_features}: \emph{Pixel} - which refers to the default \ac{MAE} proposed by He et al. \cite{mae}, and three different settings which refer to the modality used for calculating \ac{HOG} features: \emph{RGB}, \emph{Gray}, \emph{Binarized} (Bin.). While for both datasets, the largest gain of performance is reported for switching from pixel values to \ac{HOG} features, \ac{SAGHOG} works best when using HOG features of the binarized image as target on HisFrag20. All of our experiments outperform training from scratch, showing the effectiveness of pretraining.

\begin{table}
    \caption{Decoder target features.}\label{tab:tar_features}
    \tablestyle{5pt}{1.05}
    \begin{tabular}{lcc}
    \toprule
    Target feature  & Historical-WI & HisFrag20 \\ \midrule
    \demph{From scratch} & \demph{71.1} & \demph{47.4} \\ \midrule
    Pixel (RGB) & 71.9 & 50.6 \\
    \ac{HOG} (RGB) & \textbf{73.3} & 53.1 \\
    \ac{HOG} (Gray) & \textbf{73.3} & 54.1\\
    \ac{HOG} (Bin.) & \textbf{73.3} & \textbf{54.6} \\ 
    \bottomrule
    \end{tabular} 
\end{table}


\begin{table}
\caption{Ablation studies of \ac{SAGHOG} on Historical-WI (HW) and HisFrag20 (HF) with finetuned NetRVLAD. Default settings are marked in \colorbox{defaultcolor}{gray}.}\label{tab:ablation_table}
\centering
\begin{subtable}[t]{0.3\textwidth}
\caption{Masking ratio}
\tablestyle{5pt}{1.05}
\begin{tabular}{lcc}
\toprule
~   & \multicolumn{2}{c}{mAP} \\ 
Ratio   & HW & HF \\ \midrule
45 & 72.2 & 54.4 \\
60 & 72.1 & 53.6 \\
75 & \textbf{73.3}\cellcolor{defaultcolor}  & \textbf{54.6} \cellcolor{defaultcolor}\\
90 & 72.1  & 44.6 \\
\bottomrule
\end{tabular}
\end{subtable}
\hspace{\fill}
\begin{subtable}[t]{0.3\textwidth}
\caption{Encoder dimension}\label{tab:ablation_encoder}
\tablestyle{5pt}{1.05}
\begin{tabular}{lcc}
\toprule
~   & \multicolumn{2}{c}{mAP} \\ 
Dim.   & HW & HF \\ \midrule
128 & 71.3 & 47.6 \\
256 & 72.0 & 51.7\\
384 & 72.5 & 49.1 \\
512 & \textbf{73.3} \cellcolor{defaultcolor}& \textbf{54.6} \cellcolor{defaultcolor} \\
\bottomrule
\end{tabular}
\end{subtable}
\hspace{\fill}
\begin{subtable}[t]{0.35\textwidth}
\caption{Binarization algorithm}
\tablestyle{5pt}{1.05}
\begin{tabular}{lcc}
\toprule
~   & \multicolumn{2}{c}{mAP} \\ 
Method   & HW & HF \\ \midrule
Otsu {\scriptsize\cite{Otsu}} & 73.0 & 54.1 \\
Sauvola {\scriptsize\cite{Sauvola}} & \textbf{73.3}  \cellcolor{defaultcolor}  & \textbf{54.6} \cellcolor{defaultcolor}  \\
Su {\scriptsize\cite{Su}} & 72.5 & 53.6 \\
\bottomrule
\end{tabular}
\label{tab:vid-mask-ratio}
\end{subtable}

\end{table}

We provide results for the critical parts of our pipeline in Table~\ref{tab:ablation_table} when finetuning on Historical-WI and HisFrag20: the ratio of the masked patches, indicating the difficulty of the reconstruction task, the dimension of the encoder and the binarization algorithm used for patch sampling and calculating HOG features. We summarize our findings in the following:

\textit{Masking ratio} \ac{SAGHOG} performs best when using a \emph{masking ratio} of 75~\%. This is higher than values reported in related work (\cite{mae}: 60~\%, \cite{maskfeat}: 40~\%). We argue that this is mostly related to the fact that the background is not considered for the reconstruction task since we are only focusing on the HOG features of the handwriting. We observe a drop in performance when setting the masking ratio too high (90~\%). \\
\indent \textit{Embedding dimension} The performance increases with higher \emph{encoder dimensions}. The highest performance is obtained for 512 with a \ac{mAP} of 73.3~\% and 54.6~\%. Note that the encoder dimension increases the network size, we assume that higher dimensions would further increase the performance, but this introduces an additional computational burden. \\
\indent \textit{Binarization} Finally, the choice of \emph{binarization algorithm} is the least influencing factor. Even global binarization algorithms such as Otsu do perform well and close to our best one, Sauvola. We think this is because the data used does not contain heavy degradation, decreasing the difficulty of the task.  

\subsection{Finetuning}

In this section, we assess two finetuning strategies: the conventional supervised approach (Sup.) and the unsupervised method (Cl-S \cite{unsupervised_icdar17}). While supervised finetuning is the commonplace default in various computer vision tasks, this shifts for \ac{WR}, where the benchmark is unsupervised training on pseudolabels, generated by the clustering of the corresponding SIFT descriptors. In Table~\ref{tab:finetuning}, the results for both finetuning strategies are shown. For a fair comparison, we also train ResNet56 and the \ac{ViT} architecture \ac{ViT}-4/8 from scratch for both settings to show the influence of \ac{SAGHOG}. 

\begin{table}
\caption{Finetuning NetRVLAD with different backbones and target labels: Supervised (\emph{Sup.}) and Cl-S \cite{unsupervised_icdar17}. We report \ac{mAP} in \%.} 
\label{tab:finetuning}
\centering
\vspace{5pt}
\tablestyle{5pt}{1.05}
\begin{tabular}{lcccc}
\toprule
 ~ & \multicolumn{2}{c}{Historical-WI} & \multicolumn{2}{c}{HisFrag20}  \\ 
~ & Sup. & Cl-S & Sup. & Cl-S  \\ \midrule

ResNet56 & \textbf{67.2} & \textbf{73.3} & 42.5  & 41.2 \\ \midrule
\ac{ViT}-4/8 \demph{(from scratch)} & 60.0 & 71.1  & 43.4 & 47.4 \\ \midrule
\textsc{Saghog} \demph{(ours)} \\
~ ~ {\small frozen backbone} & 64.7 & 64.0 & \textbf{56.3} & \textbf{57.2}  \\ 
~ ~ {\small finetuned} & 61.4 & \textbf{73.3}  & 50.7  & 54.6  \\ \midrule 
~ ~ $\mathbf{\Delta}$ to best & \mrel{2.5} & \same{0.0} &  \prel{12.9} & \prel{9.8} \\ 
\bottomrule
\end{tabular}
\end{table}

\paragraph{Historical-WI} In contrast to \cite{unsupervised_icdar17}, we have a surprisingly good baseline when training ResNet56 with NetRVLAD on writer labels (67.2~\% mAP). This setting even outperforms \ac{SAGHOG} with supervised finetuning, indicating that the pretraining task is contrary to the finetuning - freezing the backbone also improves the \ac{mAP}. For unsupervised training, \ac{SAGHOG} is on par with ResNet56 and outperforming training \ac{ViT} from scratch. We think this is mainly due to the dataset only containing pure handwriting, where a simple CNN is enough to achieve good performance. The approach seems to be limited by the finetuning task (the pseudolabels generated by Cl-S). 

\paragraph{HisFrag20} However, the application of \ac{SAGHOG} on HisFrag20, a more complex and larger dataset, drastically shifts the results towards \ac{SAGHOG}. Firstly, training \ac{ViT} on Cl-S, even from scratch, outperforms the current state of the art. Furthermore, \acp{ViT} outperforms the residual baseline in all experiments. Our best result is achieved by freezing the encoder of \ac{SAGHOG} and only training NetRVLAD, achieving 56.3~\% for supervised and 57.2~\% for unsupervised training, which is a margin of 12.9~\%/9.8~\% to the best setting when training from scratch. In conclusion, we argue that \ac{SAGHOG} works best for large, complex datasets, where networks trained from scratch fail to perform due to noisy data - either on image- or label-level - or due to a domain shift between training and test data.

\subsection{Comparison to state of the art}
We compare the results of \ac{SAGHOG} to state of the art. In addition to the previous evaluation, we extend our analysis to include a comparison of the GRK-Papyri dataset. All results are shown in Table~\ref{tab:sota}.

\begin{table}[t!]
\caption{Comparison to state of the art. ($\dagger$) denotes our own implementation, subscript $_i$ the dimension of the final page descriptor.}\label{tab:sota}
\begin{subtable}[t]{0.99\textwidth}
\centering
\caption{Historical-WI}\label{tab:historical-wi}
\tablestyle{5pt}{1.05}
\begin{tabular}{lcc}
\toprule
Method   & \ac{mAP} & Top-1 \\ \midrule
{ Cl-S + ResNet56\phantom{.} + NetRVLAD$_{512}$} {\scriptsize\cite{peer_netrvlad}} & 73.4 & 88.5 \\ 
{ Cl-S + ResNet20\phantom{.} + mVLAD$_{400}$} {\scriptsize\cite{unsupervised_icdar17}} & 73.2 & 87.6 \\
{ Cl-S + ResNet20\phantom{.} + mVLAD$_{32000}$} {\scriptsize\cite{unsupervised_icdar17}} & 74.8 & 88.6 \\
{ Pathlet/SIFT \hspace{0.47cm} + bVLAD$_{1000}$} {\scriptsize\cite{bVLAD}} & \textbf{77.1} & \textbf{90.1} \\ \midrule
{ Cl-S + \ac{SAGHOG}\phantom{..}\hspace{0.075cm} + NetRVLAD$_{512}$} & 73.3 & 88.2\\ \midrule
~ with SGR$_{k=2}^{\dagger}$ {\scriptsize\cite{peer_netrvlad}} & \textbf{80.3} & \textbf{91.1} \\
\bottomrule
\end{tabular}
\end{subtable}
\centering
\begin{subtable}[t]{0.99\textwidth}
\caption{GRK-Papyri}\label{tab:grk50}
\tablestyle{5pt}{1.05}
\begin{tabular}{lcc}
\toprule
{ Method}   & { \ac{mAP}} & { Top-1} \\ \midrule
{ Cl-S + ResNet20 \hspace{0.05cm} + mVLAD$_{32000}$} {\scriptsize\cite{christlein_papyri}}& 42.2 & 52.0 \\
{ R-SIFT \hspace{1.4cm} + mVLAD$_{32000}$} {\scriptsize\cite{christlein_papyri}} & \textbf{42.8} & 48.0 \\
{ Cl-S + ResNet20 \hspace{0.05cm} + NetRVLAD$_{2048}^{\dagger}$} {\scriptsize\cite{peer_netrvlad}} & 38.8 & 56.0 \\ \midrule
{ Cl-S + \ac{SAGHOG}\phantom{--.}\hspace{0.075cm} + NetRVLAD$_{2048}$}& 38.2 & \textbf{58.0}\\ \midrule
~ with SGR$_{k=3}^{\dagger}$ {\scriptsize\cite{peer_netrvlad}} & \textbf{47.6} & \textbf{58.0} \\

\bottomrule

\end{tabular}
\end{subtable}

\centering
\begin{subtable}[t]{0.99\textwidth}
\caption{HisFrag20}\label{tab:hisfrag20}
\tablestyle{5pt}{1.05}
\begin{tabular}{lcccc}
\toprule
~ & \multicolumn{2}{c}{Writer Retrieval} & \multicolumn{2}{c}{Page Retrieval} \\
Method   & \ac{mAP} & Top-1 & \ac{mAP} & Top-1\\ \midrule
Cl-S + mVLAD$_{3200}$ + ESVM {\scriptsize\cite{chammas}} & 33.7 & 68.9 & - & - \\
FragNet$_{512}$ {\scriptsize\cite{hisfrag20}} & 33.5 & 77.1 & 22.6 & 36.4 \\
ResNet50 + NetVLAD$_{6400}$ + ESVM {\scriptsize\cite{Chammas_2022}} & 38.1 & 77.9 & - & - \\
Cl-S + ResNet56 + NetRVLAD$_{512}^{\dagger}$ {\scriptsize\cite{peer_netrvlad}} & 41.2 & 68.5 & 23.5 & 33.3 \\ 
Feature Mixer$_{256}$ {\scriptsize\cite{peer_fm}}  & 44.0 & 81.9 &29.3 & 45.0 \\  
A-VLAD$_{500}$ {\scriptsize\cite{avlad}} & 46.6 & \textbf{85.2} & - & - \\ \midrule
Cl-S + \textsc{Saghog} + NetRVLAD$_{512}$ & \textbf{57.2} & 81.7 & \textbf{35.2} & \textbf{47.6}\\ \midrule
~ with SGR$_{k=2}^{\dagger}$ {\scriptsize\cite{peer_netrvlad}} & \textbf{68.7} & \textbf{88.7} & \textbf{54.5} & \textbf{70.6}\\
\bottomrule
\end{tabular}
\end{subtable}
\end{table}

\paragraph{Historical-WI} As discussed in the previous section, self-supervised pretraining has a limited effect on the Historical-WI dataset - probably due to the dataset containing only (binarized) handwriting. However, \ac{SAGHOG} is on par with all deep-learning-based approaches (Table~\ref{tab:historical-wi}) and does not harm the performance. Interestingly, the best approach in terms of \ac{mAP} (without considering reranking) is still relying on handcrafted features \cite{bVLAD}. 

\paragraph{GRK-Papyri} We also finetune \textsc{Saghog} on GRK-Papyri in an unsupervised manner, a small dataset of only 50 documents. The results are shown in Table~\ref{tab:grk50}. The global page descriptors used are averaged on five different models trained, similar to the ensemble method used by Christlein et al. \cite{christlein_papyri}. We also choose to decrease the dimension of \ac{SAGHOG} to 128 instead of 512 for a fair comparison. While NetVLAD is trailing the traditional \ac{VLAD} on mAP for both networks, ResNet20 and \ac{SAGHOG}, the end-to-end learned method achieves better Top-1 accuracy, with \ac{SAGHOG} outperforming related work with 58.0~\%.

\begin{figure}[t]
    \centering
    \includegraphics[width=0.95\textwidth]{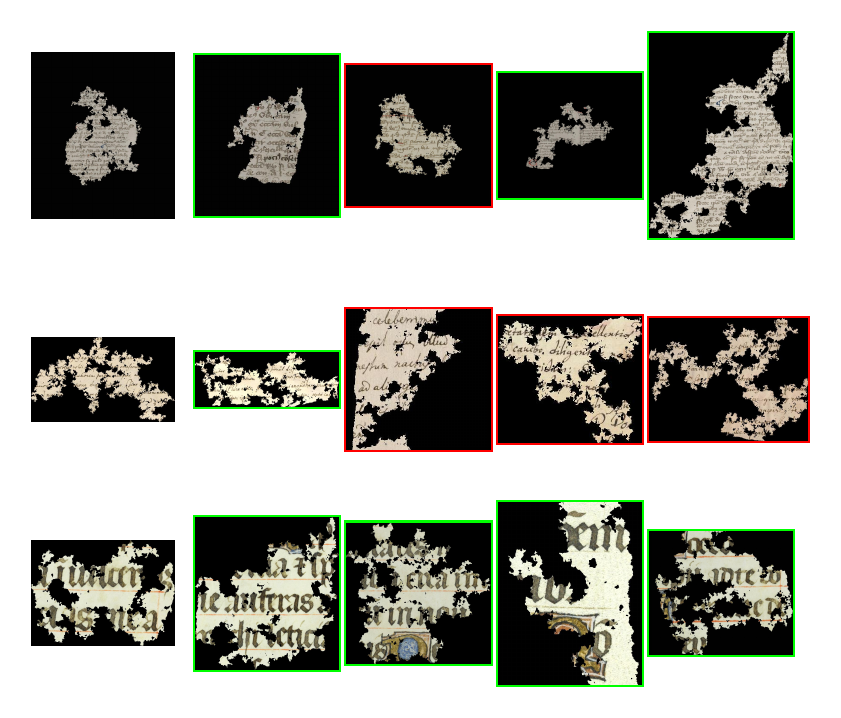}
    \caption{Qualitative results of the retrieval on HisFrag20. Left: Query. We show the four nearest documents. If the retrieved document is written by the same author as the query, we highlight the image in green, otherwise red.}
    \label{fig:qual_results}
\end{figure}

\paragraph{HisFrag20} On the largest and most diverse dataset, HisFrag20, we show that \textsc{Saghog} is superior to all previous approaches by a large margin on both tasks - \ac{WR} (+10.6~\%) and page retrieval (+5.9~\%). A thorough comparison to related work is given in Table~\ref{tab:hisfrag20}. Although the \ac{mAP} and therefore the ranked list is better, \ac{SAGHOG} has a slightly worse Top-1 accuracy. As we showed in our work introducing the feature mixer \cite{peer_fm}, networks trained on full-color images tend to infer features from the background during training, which is also shown in our comparison: For example, approaches relying on $32\times32$ patches such as \cite{chammas,peer_netrvlad} suffer from a limited Top-1 accuracy. We think this is due to approaches on image-level retrieving images from the same page, where the focus is not on retrieving documents based on the similarity of the handwriting, but rather on background similarity, which is shared within a page. This phenomenon is also observed by considering the page retrieval performance (e.g., \cite{peer_netrvlad} vs. \cite{peer_fm}). However, we argue that due to \ac{SAGHOG} achieving a superior performance in terms of \ac{mAP}, we can apply a reranking scheme such as \ac{SGR} which is shown to be beneficial in particular for Top-1 accuracy \cite{peer_netrvlad}. By using \ac{SGR}, we are able to boost both metrics by 11.5~\% \ac{mAP} / 7~\% Top-1 for \ac{WR} and 19.3~\% \ac{mAP} / 23~\% Top-1 for page retrieval, which outperforms related work, even on the Top-1 accuracy. We also provide qualitative results in Fig.~\ref{fig:qual_results}.

\section{Conclusion}\label{sec:conclusion}
In this paper, we introduce \ac{SAGHOG}, the first masking-based self-supervised approach for the downstream task of \ac{WR}. Our pretraining includes reconstructing \ac{HOG} features of the binarized image, which leads the network to focus on handwriting and suppress the background. Additionally, we propose a preprocessing via \ac{SAM} to make use of previous historical documents for our pretraining. We evaluate three different datasets and two different finetuning methods commonly used in the domain of \ac{WR}. In the following, we summarize our key results and ideas for future work.
\paragraph{Our findings} Firstly, we investigate our preprocessing scheme and show that the segmentation of handwriting is beneficial for the pretraining of \ac{SAGHOG}. Secondly, \ac{SAGHOG} shines on \ac{WR} for large, complex datasets containing sources of noise (e.g. HisFrag20 with small, torn fragments). We are able to outperform related work by a significant margin on this difficult dataset. Furthermore, we show that for a binarized dataset such as Historical-WI the effect of pretraining is limited, for which current methods are already working decent. In the end, we also evaluate on GRK-Papyri, a small dataset, and show that the application of \ac{SAGHOG} improves the Top-1 accuracy upon state of the art. \ac{SAGHOG} also shows the applicability of the \ac{ViT} architecture for \ac{WR}, which heavily depends on CNNs.
\paragraph{Future work} We further want to investigate self-supervised learning for. On the one hand, one could improve the diversity of data for pretraining (we mainly use Latin script and documents including a limited amount of degradation) to improve results also on smaller datasets such as GRK-Papyri. On the other hand, tuning the reconstruction task, e.g., using a U-Net for improved binarization or integrating style embeddings from specific domains to refine the self-supervised embeddings, promises to further improve the results.

\paragraph{Acknowledgements}
We thank Vincent Christlein for providing the binarized images of GRK-Papyri.

\bibliographystyle{splncs04}
\bibliography{bib}

\begin{thebibliography}{10}
\providecommand{\url}[1]{\texttt{#1}}
\providecommand{\urlprefix}{URL }
\providecommand{\doi}[1]{https://doi.org/#1}

\bibitem{dino}
Caron, M., Touvron, H., Misra, I., J{\'{e}}gou, H., Mairal, J., Bojanowski, P., Joulin, A.: Emerging properties in self-supervised vision transformers. In: 2021 {IEEE/CVF} International Conference on Computer Vision, {ICCV} 2021, Montreal, QC, Canada, October 10-17, 2021. pp. 9630--9640 (2021)

\bibitem{chammas}
Chammas, M., Makhoul, A., Demerjian, J.: Writer identification for historical handwritten documents using a single feature extraction method. In: 19th {IEEE} International Conference on Machine Learning and Applications, {ICMLA} 2020, Miami, FL, USA, December 14-17, 2020. pp.~1--6 (2020)

\bibitem{Chammas_2022}
Chammas, M., Makhoul, A., Demerjian, J., Dannaoui, E.: A deep learning based system for writer identification in handwritten arabic historical manuscripts. Multimedia Tools and Applications  \textbf{81}(21),  30769–30784 (2022)

\bibitem{simclr}
Chen, T., Kornblith, S., Norouzi, M., Hinton, G.E.: A simple framework for contrastive learning of visual representations. In: Proceedings of the 37th International Conference on Machine Learning, {ICML} 2020, 13-18 July 2020, Virtual Event. Proceedings of Machine Learning Research, vol.~119, pp. 1597--1607 (2020)

\bibitem{zernike}
Christlein, V., Bernecker, D., Angelopoulou, E.: Writer identification using {VLAD} encoded contour-zernike moments. In: 13th International Conference on Document Analysis and Recognition, {ICDAR} 2015, Nancy, France, August 23-26, 2015. pp. 906--910 (2015)

\bibitem{unsupervised_icdar17}
Christlein, V., Gropp, M., Fiel, S., Maier, A.K.: Unsupervised feature learning for writer identification and writer retrieval. In: 14th {IAPR} International Conference on Document Analysis and Recognition, {ICDAR} 2017, Kyoto, Japan, November 9-15, 2017. pp. 991--997 (2017)

\bibitem{christlein_cnn_vlad}
Christlein, V., Maier, A.K.: Encoding {CNN} activations for writer recognition. In: 13th {IAPR} International Workshop on Document Analysis Systems, {DAS} 2018, Vienna, Austria, April 24-27, 2018. pp. 169--174 (2018)

\bibitem{christlein_papyri}
Christlein, V., Marthot{-}Santaniello, I., Mayr, M., Nicolaou, A., Seuret, M.: Writer retrieval and writer identification in greek papyri. In: Intertwining Graphonomics with Human Movements - 20th International Conference of the International Graphonomics Society, {IGS} 2021, Las Palmas de Gran Canaria, Spain, June 7-9, 2022, Proceedings. vol. 13424, pp. 76--89 (2022)

\bibitem{icdar19}
Christlein, V., Nicolaou, A., Seuret, M., Stutzmann, D., Maier, A.: {ICDAR} 2019 competition on image retrieval for historical handwritten documents. In: 2019 International Conference on Document Analysis and Recognition, {ICDAR} 2019, Sydney, Australia, September 20-25, 2019. pp. 1505--1509 (2019)

\bibitem{clamm2017}
Cloppet, F., Eglin, V., Helias{-}Baron, M., Kieu, V.C., Vincent, N., Stutzmann, D.: {ICDAR2017} competition on the classification of medieval handwritings in latin script. In: 14th {IAPR} International Conference on Document Analysis and Recognition, {ICDAR} 2017, Kyoto, Japan, November 9-15, 2017. pp. 1371--1376 (2017)

\bibitem{clamm2016}
Cloppet, F., Eglin, V., Kieu, V.C., Stutzmann, D., Vincent, N.: {ICFHR2016} competition on the classification of medieval handwritings in latin script. In: 15th International Conference on Frontiers in Handwriting Recognition, {ICFHR} 2016, Shenzhen, China, October 23-26, 2016. pp. 590--595 (2016)

\bibitem{cbad}
Diem, M., Kleber, F., Sablatnig, R., Gatos, B.: {cBAD}: {ICDAR2019} competition on baseline detection. In: 2019 International Conference on Document Analysis and Recognition, {ICDAR} 2019, Sydney, Australia, September 20-25, 2019. pp. 1494--1498 (2019)

\bibitem{vit}
Dosovitskiy, A., Beyer, L., Kolesnikov, A., Weissenborn, D., Zhai, X., Unterthiner, T., Dehghani, M., Minderer, M., Heigold, G., Gelly, S., Uszkoreit, J., Houlsby, N.: An image is worth 16x16 words: Transformers for image recognition at scale. In: 9th International Conference on Learning Representations, {ICLR} 2021, Virtual Event, Austria, May 3-7, 2021 (2021)

\bibitem{icdar2017}
Fiel, S., Kleber, F., Diem, M., Christlein, V., Louloudis, G., Nikos, S., Gatos, B.: {ICDAR2017} competition on historical document writer identification (historical-wi). In: 14th {IAPR} International Conference on Document Analysis and Recognition, {ICDAR} 2017, Kyoto, Japan, November 9-15, 2017. pp. 1377--1382 (2017)

\bibitem{mae}
He, K., Chen, X., Xie, S., Li, Y., Doll{\'{a}}r, P., Girshick, R.B.: Masked autoencoders are scalable vision learners. In: {IEEE/CVF} Conference on Computer Vision and Pattern Recognition, {CVPR} 2022, New Orleans, LA, USA, June 18-24, 2022. pp. 15979--15988. {IEEE} (2022)

\bibitem{moco}
He, K., Fan, H., Wu, Y., Xie, S., Girshick, R.B.: Momentum contrast for unsupervised visual representation learning. In: 2020 {IEEE/CVF} Conference on Computer Vision and Pattern Recognition, {CVPR} 2020, Seattle, WA, USA, June 13-19, 2020. pp. 9726--9735. Computer Vision Foundation / {IEEE} (2020)

\bibitem{visual_burstiness}
J{\'{e}}gou, H., Douze, M., Schmid, C.: On the burstiness of visual elements. In: 2009 {IEEE} Computer Society Conference on Computer Vision and Pattern Recognition {(CVPR} 2009), 20-25 June 2009, Miami, Florida, {USA}. pp. 1169--1176 (2009)

\bibitem{keglevic}
Keglevic, M., Fiel, S., Sablatnig, R.: Learning features for writer retrieval and identification using triplet cnns. In: 16th International Conference on Frontiers in Handwriting Recognition, {ICFHR} 2018, Niagara Falls, NY, USA, August 5-8, 2018. pp. 211--216 (2018)

\bibitem{adam}
Kingma, D.P., Ba, J.: Adam: {A} method for stochastic optimization. In: Bengio, Y., LeCun, Y. (eds.) 3rd International Conference on Learning Representations, {ICLR} 2015, San Diego, CA, USA, May 7-9, 2015, Conference Track Proceedings (2015)

\bibitem{sam}
Kirillov, A., Mintun, E., Ravi, N., Mao, H., Rolland, C., Gustafson, L., Xiao, T., Whitehead, S., Berg, A.C., Lo, W.Y., Dollar, P., Girshick, R.: Segment anything. In: Proceedings of the IEEE/CVF International Conference on Computer Vision (ICCV). pp. 4015--4026 (2023)

\bibitem{bVLAD}
Lai, S., Zhu, Y., Jin, L.: Encoding pathlet and {SIFT} features with bagged {VLAD} for historical writer identification. {IEEE} Trans. Inf. Forensics Secur.  \textbf{15},  3553--3566 (2020)

\bibitem{Lastilla2022}
Lastilla, L., Ammirati, S., Firmani, D., Komodakis, N., Merialdo, P., Scardapane, S.: Self-supervised learning for medieval handwriting identification: A case study from the vatican apostolic library. Information Processing &amp; Management  \textbf{59}(3),  102875 (2022)

\bibitem{cosine_decay}
Loshchilov, I., Hutter, F.: {SGDR:} stochastic gradient descent with warm restarts. In: 5th International Conference on Learning Representations, {ICLR} 2017, Toulon, France, April 24-26, 2017, Conference Track Proceedings. OpenReview.net (2017)

\bibitem{adamw}
Loshchilov, I., Hutter, F.: Decoupled weight decay regularization. In: 7th International Conference on Learning Representations, {ICLR} 2019, New Orleans, LA, USA, May 6-9, 2019 (2019)

\bibitem{grk-papyri}
Mohammed, H.A., Marthot{-}Santaniello, I., M{\"{a}}rgner, V.: Grk-papyri: {A} dataset of greek handwriting on papyri for the task of writer identification. In: 2019 International Conference on Document Analysis and Recognition, {ICDAR} 2019, Sydney, Australia, September 20-25, 2019. pp. 726--731 (2019)

\bibitem{avlad}
Ngo, T.T., Nguyen, H.T., Nakagawa, M.: {A-VLAD:} an end-to-end attention-based neural network for writer identification in historical documents. In: 16th International Conference on Document Analysis and Recognition, {ICDAR} 2021, Lausanne, Switzerland, September 5-10, 2021, Proceedings, Part {II}. vol. 12822, pp. 396--409 (2021)

\bibitem{Otsu}
Otsu, N.: A threshold selection method from gray-level histograms. IEEE Transactions on Systems, Man, and Cybernetics  \textbf{9}(1),  62–66 (Jan 1979)

\bibitem{peer_selfsupervised}
Peer, M., Kleber, F., Sablatnig, R.: Self-supervised vision transformers with data augmentation strategies using morphological operations for writer retrieval. In: Frontiers in Handwriting Recognition - 18th International Conference, {ICFHR} 2022, Hyderabad, India, December 4-7, 2022, Proceedings. pp. 122--136 (2022)

\bibitem{peer_netrvlad}
Peer, M., Kleber, F., Sablatnig, R.: Towards writer retrieval for historical datasets. In: Document Analysis and Recognition - {ICDAR} 2023 - 17th International Conference, San Jos{\'{e}}, CA, USA, August 21-26, 2023, Proceedings, Part {I}. pp. 411--427 (2023)

\bibitem{peer_fm}
Peer, M., Sablatnig, R.: Feature mixing for writer retrieval and identification on papyri fragments. In: Proceedings of the 7th International Workshop on Historical Document Imaging and Processing (2023)

\bibitem{Sauvola}
Sauvola, J., Pietik\"{a}inen, M.: Adaptive document image binarization. Pattern Recognition  \textbf{33}(2),  225–236 (Feb 2000)

\bibitem{hisfrag20}
Seuret, M., Nicolaou, A., Maier, A., Christlein, V., Stutzmann, D.: {ICFHR} 2020 competition on image retrieval for historical handwritten fragments. In: 17th International Conference on Frontiers in Handwriting Recognition, {ICFHR} 2020, Dortmund, Germany, September 8-10, 2020. pp. 216--221 (2020)

\bibitem{Su}
Su, B., Lu, S., Tan, C.L.: Binarization of historical document images using the local maximum and minimum. In: Proceedings of the 9th IAPR International Workshop on Document Analysis Systems (Jun 2010)

\bibitem{msloss}
Wang, X., Han, X., Huang, W., Dong, D., Scott, M.R.: Multi-similarity loss with general pair weighting for deep metric learning. In: {IEEE} Conference on Computer Vision and Pattern Recognition, {CVPR} 2019, Long Beach, CA, USA, June 16-20, 2019. pp. 5022--5030 (2019)

\bibitem{wang_supervised}
Wang, Z., Maier, A., Christlein, V.: Towards end-to-end deep learning-based writer identification. In: 50. Jahrestagung der Gesellschaft f{\"{u}}r Informatik, {INFORMATIK} 2020 - Back to the Future, Karlsruhe, Germany, 28. September - 2. Oktober 2020. vol. {P-307}, pp. 1345--1354 (2020)

\bibitem{maskfeat}
Wei, C., Fan, H., Xie, S., Wu, C., Yuille, A.L., Feichtenhofer, C.: Masked feature prediction for self-supervised visual pre-training. In: {IEEE/CVF} Conference on Computer Vision and Pattern Recognition, {CVPR} 2022, New Orleans, LA, USA, June 18-24, 2022. pp. 14648--14658 (2022)

\bibitem{Zenk2023}
Zenk, J., Kordon, F., Mayr, M., Seuret, M., Christlein, V.: Investigations on self-supervised learning for script-, font-type, and location classification on historical documents. In: Proceedings of the 7th International Workshop on Historical Document Imaging and Processing. HIP ’23 (Aug 2023)

\end{thebibliography}
\end{document}